# Fine-Tuned CNN-Based Approach for Multi-Class Mango Leaf Disease Detection

Jalal Ahmmed
*Department of Computer Science and Engineering*
*Daffodil International University*
Dhaka, Bangladesh
ahmmed22205101088@diu.edu.bd

Faruk Ahmed
*Department of Computer Science and Engineering*
*Daffodil International University*
Dhaka, Bangladesh
faruk15-4205@diu.edu.bd

Rashedul Hasan Shohan
*Department of Computer Science and Engineering*
*Daffodil International University*
Dhaka, Bangladesh
shohan15-6526@s.diu.edu.bd

Md.Mahabub Rana
*Department of Computer Science and Engineering*
*Daffodil International University*
Dhaka, Bangladesh
rana22205101234@diu.edu.bd

Mahdi Hasan
*Department of Computer Science and Engineering*
*Daffodil International University*
Dhaka, Bangladesh
hasan22205101035@diu.edu.bd

*Abstract*—Mango is an important fruit crop in South Asia, but its cultivation is frequently hampered by leaf diseases that greatly impact yield and quality. This research examines the performance of five pre-trained convolutional neural networks, DenseNet201, InceptionV3, ResNet152V2, SeResNet152, and Xception, for multi-class identification of mango leaf diseases across eight classes using a transfer learning strategy with fine-tuning. The models were assessed through standard evaluation metrics, such as accuracy, precision, recall, F1-score, and confusion matrices. Among the architectures tested, DenseNet201 delivered the best results, achieving 99.33% accuracy with consistently strong metrics for individual classes, particularly excelling in identifying Cutting Weevil and Bacterial Canker. Moreover, ResNet152V2 and SeResNet152 provided strong outcomes, whereas InceptionV3 and Xception exhibited lower performance in visually similar categories like Sooty Mould and Powdery Mildew. The training and validation plots demonstrated stable convergence for the highest-performing models. The capability of fine-tuned transfer learning models, for precise and dependable multi-class mango leaf disease detection in intelligent agricultural applications.

## I. INTRODUCTION

Mango (*Mangifera indica*) is one of the most economically significant fruits cultivated in South Asia, particularly in countries like Bangladesh and India. It contributes greatly to the livelihood of farmers and plays a crucial role in strengthening the agricultural economy of the region [1]. Despite its importance, mango cultivation is frequently threatened by numerous leaf diseases, including anthracnose, bacterial canker, and powdery mildew. These diseases can cause serious damage to both yield quality and quantity, resulting in notable financial setbacks for farmers [2]. Traditionally, the detection of mango leaf diseases has relied on manual inspection by farmers and agricultural professionals. However, this method is highly subjective and depends on visual interpretation, which often leads to misdiagnosis. Furthermore, it is inefficient and labor-intensive when monitoring large-scale orchards [3].

To mitigate these issues, the use of automated disease recognition systems has gained importance in modern agriculture. These systems can significantly speed up the diagnosis process and help prevent crop loss by enabling early detection [4]. Among emerging digital agricultural technologies, image-based deep learning techniques have demonstrated strong potential. Specifically, Convolutional Neural Networks (CNNs) have demonstrated excellent performance in classifying plant diseases from leaf images [5]. CNNs are favored due to their capacity to automatically extract intricate features from raw image data, eliminating the need for manually designed features [6]. However, building deep CNNs from the ground up requires substantial labeled datasets and significant computational power, which can be challenging to obtain in developing nations like Bangladesh [3][7].

To address these challenges, transfer learning has proven to be an effective alternative [8]. This method allows researchers to adapt models pre-trained on large-scale datasets like ImageNet to specific agricultural tasks [9][10]. As a result, high-performance models can be developed even with relatively small agricultural datasets, while also minimizing training time and resource consumption [11].

In this research, five powerful CNN architectures, DenseNet201, InceptionV3, ResNet152V2, SeResNet152, and Xception, are fine-tuned and evaluated for automated classification of eight mango leaf disease categories. The focus is on identifying the most effective architecture based on empirical performance in a controlled training and testing pipeline.

The main contributions of this study are:

- A transfer learning-based classification system was developed using five fine-tuned CNN models, enhanced with layers like GlobalAveragePooling2D, BatchNormalization, Dropout, and Dense for improved feature learning.

- Unlike many prior studies that emphasize only accuracy, this work presents a comprehensive multi-metric evaluation, highlighting how each model performs in terms of precision, recall, and F1-score for each disease class. This thorough metric-based analysis ensures a more realistic and balanced assessment of model reliability, especially in class-sensitive agricultural applications.

- DenseNet201 achieved the best overall performance, with a classification accuracy of 99.33%, and showed consistently high precision (up to 100%), recall (up to 100%), and F1-scores (up to 100%) across eight mango leaf disease classes. Its robust performance, particularly in classes like Cutting Weevil and Die

Back, confirmed its suitability for practical deployment.

## II. RELATED WORKS

Recent developments in plant disease identification have primarily aimed to build lightweight, high-accuracy models while tackling challenges related to limited datasets. Zhang et al. [9] introduced CBAM-DBIRNet, a compact dual-branch network integrated with CBAM attention and depthwise separable convolutions, achieving 98.42% accuracy with a model size of just 0.64 MB for grading anthracnose severity in mango leaves. Similarly, Thanjaivadivel et al. [10] improved CNN models by incorporating depthwise convolutions and inverted residual connections, reaching 99.87% accuracy over 39 plant disease classes by incorporating features such as color and shape.

Numerous studies have applied deep learning to the agricultural domain. Shoaib et al. [11] conducted a comparative evaluation and recognized InceptionV3 as the top-performing model with 99.87% accuracy. Srivastava and Meena [12] tested several architectures, including VGG16, MobileNetV2, DenseNet201, and Xception, reporting accuracies of 98.9% and 99.9% across two distinct datasets. Pathak et al. [13] developed a CNN model for detecting mango leaf diseases and successfully deployed it in an Android app with 99% classification accuracy. To handle data scarcity, Ramadan et al. [14] utilized CycleGAN to generate synthetic samples, enhancing MobileNet's performance to 98.54%. Likewise, Jha et al. [15] proposed a stacked ensemble model combining Residual Network, MobileNet, and Inception, achieving 98.86% accuracy for potato leaf disease classification.

Fan et al. [16] introduced ensemble-based PDDNet models using early fusion and lead-voting strategies across nine pre-trained CNNs, which yielded 96.74% and 97.79% accuracy in a 15-class problem. A review by Jadhav-Mane and Singh [17] offered a comprehensive summary of mango leaf disease detection approaches, outlining research challenges and areas for future innovation. In another work, Shafik et al. [18] demonstrated that combining handcrafted and deep features through transfer learning and center loss could push accuracy up to 99.79%, underscoring the effectiveness of hybrid feature extraction.

These studies collectively reflect a growing focus on compact model architectures, ensemble techniques, synthetic data generation, and multi-feature fusion as effective strategies for boosting classification accuracy and enabling real-world agricultural deployment. Most reported systems attain accuracy levels of 97% or higher, reinforcing the potential of deep learning in plant disease diagnostics.

## III. MATERIALS AND METHODOLOGY

This study employs a deep learning strategy using transfer learning to categorize mango leaf diseases into eight distinct classes, encompassing comprehensive dataset preparation, meticulous preprocessing and augmentation techniques, strategic model selection, fine-tuning of pretrained convolutional neural networks (CNNs), and thorough assessment with a variety of performance metrics, effectively leveraging reusable visual feature representations to significantly minimize the need for extensive annotated agricultural data and substantial computational expense, with the detailed architectural workflow clearly depicted in Fig. 1.

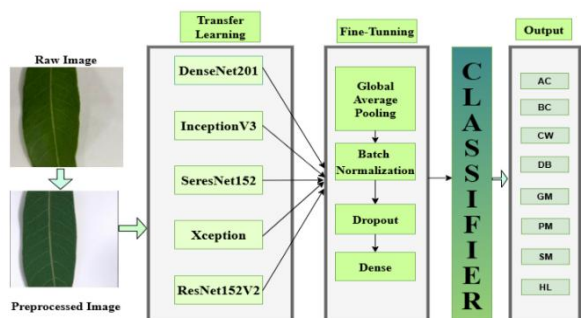

Fig. 1. Proposed Methodology

### A. Dataset Description

The dataset used in this study includes 6,400 RGB images collected from mango orchards in Kushtia and Dhaka, Bangladesh, and is publicly available on Mendeley Data. It covers eight mango leaf categories, Anthracnose (AC), Bacterial Canker (BC), Cutting Weevil (CW), Die Back (DB), Gall Midge (GM), Powdery Mildew (PM), Sooty Mould (SM), and Healthy (HL), with 800 images per category. Images, captured with an iPhone SE at 3024 × 4032 pixels, were resized to 240 × 240 pixels to enhance memory usage and computational efficiency. Fig. 2 displays sample images from each category.

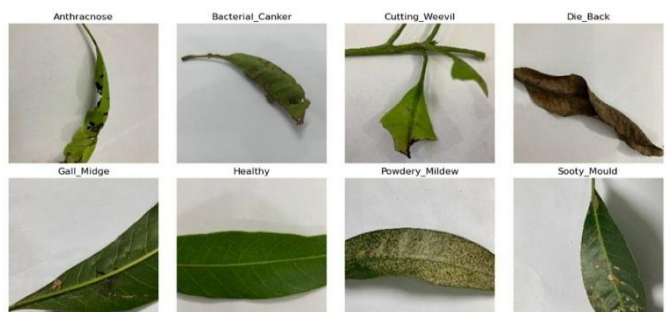

Fig. 2. Original image of different classes of Mango Leaf

### B. Data Preprocessing and Augmentation

To enhance image diversity and consistency, preprocessing and augmentation techniques were applied. Gaussian blurring reduced high-frequency noise while preserving edge features, and Contrast Limited Adaptive Histogram Equalization (CLAHE) improved contrast to highlight disease-affected regions. Images were resized to 240 × 240 pixels and normalized to a 0-1 intensity range. In Fig. 3, real-time augmentation during training, with random rotations up to 30 degrees, horizontal/vertical flips, zooming, and brightness adjustments (0.8x-1.2x), increased the training set from 6,400 to 33,600 images, with 13,200 reserved for validation/testing (80:20 ratio). Labels were encoded using one-hot encoding.

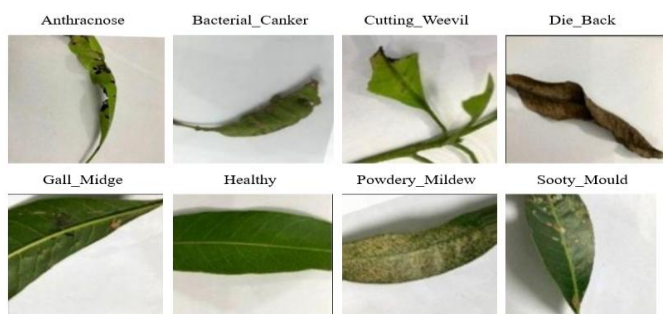

Fig. 3. Augmented image of different classes of Mango Leaf

## C. Model Selection

This study initially evaluated five popular transfer learning architectures: DenseNet201, InceptionV3, ResNet152V2, SeResNet152, and Xception. These models were selected based on their proven accuracy on the ImageNet benchmark and their architectural diversity [19]. DenseNet201 was chosen as the final focus due to its superior performance during comparative evaluation. This architecture features a dense connectivity design, where every layer obtains feature maps from all previous layers, enhancing effective gradient flow and encouraging feature reuse. DenseNet201 strikes a good balance among depth, accuracy, and computational efficiency, containing 20.2 million parameters and a relatively small model size of about 77 MB. Although InceptionV3 and Xception excel at multi-scale feature extraction, and SeResNet152 uses attention mechanisms, DenseNet201 demonstrated the most reliable performance across all evaluation metrics, making it the preferred choice for fine-tuning.

## D. Fine Tuning

In this study, fine-tuning is crucial for improving model performance by utilizing the robust feature representations of a pretrained convolutional neural network (CNN), specifically DenseNet201, trained on the large-scale ImageNet dataset. Rather than training a model from scratch, which demands extensive data and computational resources, fine-tuning allows the reuse of learned low- and mid-level features such as edges, textures, and patterns that are transferable across visual domains. By freezing the convolutional base and adding custom classification layers tailored to the specific dataset, we effectively minimize overfitting while speeding up convergence. Additionally, techniques like global average pooling, dropout regularization, batch normalization, and a task-specific dense output layer enhance the model's adaptation to the new domain with better generalization.

*a) Input and Preprocessing:* Let $X \in R^{H \times W \times C}$ denote the input image, where H, W and C represent the height, width, and number of channels, respectively. Prior to model input, images are resized, normalised, and optionally augmented to enhance generalization capability during training.

*b) Feature Extraction:* A pre-trained convolutional neural network (CNN), such as DenseNet201, acts as a feature extractor. The image X is forwarded through the base network $f_{\text{base}}(X)$ to obtain feature maps:

$$F = f_{\text{base}}(X), \quad F \in R^{h \times w \times d} \quad (1)$$

$F \in R^{H \times W \times d}$ is the intermediate feature map.

Here, h, W and d denote the height, width, and depth of the extracted features, respectively. The base network is frozen to preserve pre-trained weights.

*c) Average Pooling (GAP):* To reduce spatial dimensions and create a fixed-length representation, global average pooling is applied over the feature maps:

$$G_k = \frac{1}{h \cdot w} \sum_{i=1}^{h} \sum_{j=1}^{w} F_{i,j,k}, \quad \text{for } k = 1,2,\ldots,d \quad (2)$$

Coverts $F \in R^{h \times w \times d}$ into $G \in R^d$

Reduces spatial dimensions while preserving depth-wise features.

*d) Dense Layer with Softmax Activation:* After dropout, the feature vector is passed to a dense layer for classification:

$$Z = WD + b \quad (3)$$

Softmax is then applied to compute class probabilities:

$$\hat{y}_i = \frac{e^{Z_i}}{\sum_{j=1}^{C} e^{Z_j}}, \quad \text{for } i = 1,\ldots,C \quad (4)$$

## E. Model Training Parameters

The training process was applied uniformly to all five fine-tuned models: DenseNet201, InceptionV3, ResNet152V2, SeResNet152, and Xception. Training occurred in two phases: initially, the convolutional base (pre-trained on ImageNet) was frozen, and only the new classification layers were trained. In the second phase, the entire model was unfrozen and fine-tuned with a lower learning rate to retain general features while adapting to the mango leaf dataset. In Table 1, all models used the Adam optimizer with a base learning rate of 1e−4 and the SparseCategoricalCrossentropy loss function (from_logits=True). A batch size of 32, early stopping (patience = 10), and step-decay learning rate scheduling were applied to improve generalization and prevent overfitting. Input images were resized to 224 × 224 × 3, and training was conducted for 150 epochs on a consistent dataset split.

TABLE I. TRAINING PARAMETER

| Hyperparameter |
| --- |
| Epochs = 150 |
| Batch size = 32 |
| Image size = (224, 224, 3) |
| Learning rate = 1e−4 |
| Weight decay = 1e−7 |
| Optimizer = Adam |
| Loss function = SparseCategoricalCrossentropy(from_logits=True) |
| Early stopping = EarlyStopping(monitor='val_accuracy', patience=10, verbose=1, restore_best_weights=True) |
| Learning rate scheduler = LearningRateScheduler(lambda epoch: learning_rate *0.1 ** (epoch // 10)) |

## F. Hardware Setup

All training and evaluation tasks were performed on Google Colab, which provides free access to an NVIDIA Tesla T4 GPU with 16 GB RAM. This setup enabled efficient training of deep CNNs with moderately large datasets. No additional local or cloud-based hardware was used.

## G. Performance Evaluation Metrics

To assess model effectiveness comprehensively, four standard classification metrics were employed: Accuracy, Precision, Recall, and F1-Score. These metrics quantify overall correctness, the reliability of positive predictions, sensitivity to true positives, and the harmonic balance between precision and recall, respectively. Their mathematical formulations are:

$$\text{Accuracy} = \frac{\text{TP} + \text{TN}}{\text{TP} + \text{TN} + \text{FP} + \text{FN}} \quad (5)$$

$$\text{Precision} = \frac{\text{TP}}{\text{TP} + \text{FP}} \quad (6)$$

$$\text{Recall} = \frac{\text{TP}}{\text{TP} + \text{FN}} \quad (7)$$

$$F1 = 2 \cdot \frac{\text{Precision} \cdot \text{Recall}}{\text{Precision} + \text{Recall}} \quad (8)$$

Where TP, TN, FP, and FN denote true positives, true negatives, false positives, and false negatives, respectively. These metrics were calculated for each disease class and averaged to determine overall model performance.

## IV. RESULTS AND DISCUSSION

### A. Classification Report of Fine-tuned Models

TABLE II. PERFORMANCE METRICS OF FINE-TUNED MODELS

| Model | Class | Precision | Recall | F1-score | Accuracy |
|---|---|---|---|---|---|
| DenseNet201 | AC | 1.00 | 0.98 | 0.99 | 99.33% |
|  | BC | 0.99 | 0.99 | 0.99 |  |
|  | CW | 1.00 | 1.00 | 1.00 |  |
|  | DB | 0.99 | 1.00 | 1.00 |  |
|  | GM | 0.99 | 0.98 | 0.99 |  |
|  | HL | 0.99 | 1.00 | 0.99 |  |
|  | PM | 0.99 | 0.99 | 0.99 |  |
|  | SM | 0.99 | 0.98 | 0.98 |  |
| InceptionV3 | AC | 0.98 | 0.97 | 0.97 | 98.66% |
|  | BC | 0.98 | 0.98 | 0.98 |  |
|  | CW | 1.00 | 1.00 | 1.00 |  |
|  | DB | 0.98 | 0.99 | 0.98 |  |
|  | GM | 0.97 | 0.97 | 0.97 |  |
|  | HL | 0.98 | 0.99 | 0.98 |  |
|  | PM | 0.95 | 0.95 | 0.95 |  |
|  | SM | 0.94 | 0.93 | 0.93 |  |
| ResNet152V2 | AC | 0.99 | 0.98 | 0.99 | 99.16% |
|  | BC | 0.99 | 0.99 | 0.99 |  |
|  | CW | 1.00 | 1.00 | 1.00 |  |
|  | DB | 0.99 | 1.00 | 0.99 |  |
|  | GM | 0.97 | 0.98 | 0.98 |  |
|  | HL | 0.99 | 1.00 | 0.99 |  |
|  | PM | 0.97 | 0.98 | 0.97 |  |
|  | SM | 0.97 | 0.96 | 0.96 |  |
| SeresNet152 | AC | 0.99 | 0.99 | 0.99 | 99.16% |
|  | BC | 1.00 | 0.99 | 1.00 |  |
|  | CW | 1.00 | 1.00 | 1.00 |  |
|  | DB | 0.99 | 1.00 | 0.99 |  |
|  | GM | 0.99 | 0.98 | 0.99 |  |
|  | HL | 0.99 | 0.99 | 0.99 |  |
|  | PM | 0.99 | 0.99 | 0.99 |  |
|  | SM | 0.99 | 0.98 | 0.99 |  |
| Xception | AC | 0.98 | 0.98 | 0.98 | 98.42% |
|  | BC | 0.99 | 0.97 | 0.98 |  |
|  | CW | 1.00 | 1.00 | 1.00 |  |
|  | DB | 0.98 | 0.99 | 0.99 |  |
|  | GM | 0.98 | 0.97 | 0.97 |  |
|  | HL | 0.99 | 0.99 | 0.99 |  |
|  | PM | 0.97 | 0.95 | 0.96 |  |
|  | SM | 0.93 | 0.96 | 0.95 |  |

In Table 2, the evaluation results reveal that DenseNet201 achieved the highest classification accuracy of 99.33% among the five transfer learning models, with consistently strong precision, recall, and F1-scores across all eight classes. Specifically, DenseNet201 exhibited 100% precision for Anthracnose and Cutting Weevil and maintained 98–100% recall and F1-scores for all disease types. InceptionV3 also performed well with an accuracy of 98.66%, showing slight variations in recall and F1-scores for Sooty Mould and Healthy leaves. ResNet152V2 and SeResNet152 both achieved an accuracy of 99.16%, demonstrating particularly high precision and recall values above 97% for all classes. Xception achieved 98.42% accuracy, with precision and recall values slightly lower for categories such as Healthy and Sooty Mould.

### B. The training and validation loss curves

The training and validation loss curves of each model are illustrated in Fig. 4. DenseNet201, ResNet152V2, and SeResNet152 demonstrated stable and effective learning, with closely aligned loss curves indicating strong generalization performance. DenseNet201's training loss decreased from 1.1 to 0.04 by epoch 130, while ResNet152V2 and SeResNet152 showed similar convergence, reaching final losses around 0.03–0.04 by epochs 90 and 120, respectively. In contrast, InceptionV3 and Xception exhibited a slight divergence between training and validation losses after epoch 100, with validation losses plateauing near 0.07 and 0.06. Nevertheless, all five models achieved substantial reductions from their initial loss values, reflecting strong overall learning capacity.

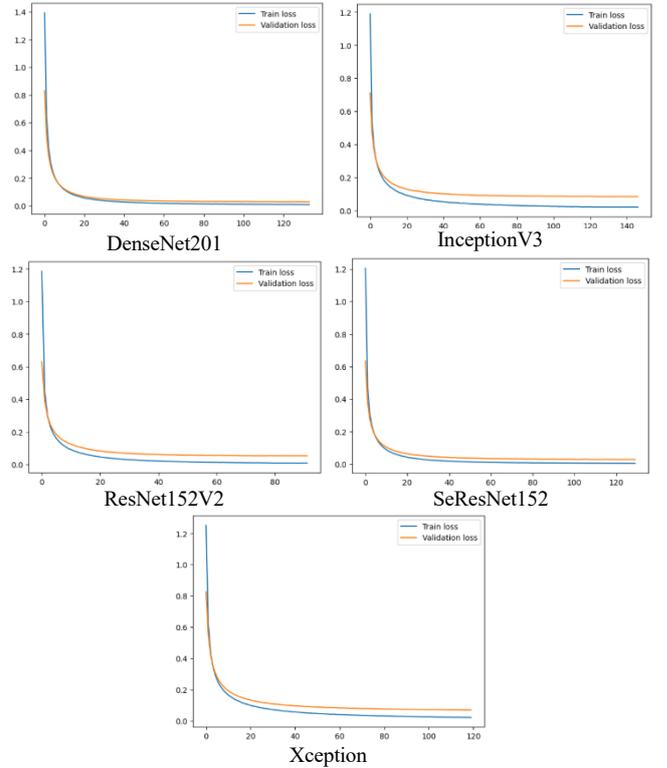

Fig. 4. Training and Validation loss curves of Fine-tuned CNNs

### C. Confusion Matrix of the models

The confusion matrices (Figs. 5–9) highlight class-wise prediction performance across the fine-tuned models. DenseNet201 (Fig. 5) demonstrated the highest class separability, achieving perfect classification for Cutting Weevil (1647/1647) and near-perfect results for Anthracnose (1621/1628) and Gall Midge (1625/1642), with very few misclassifications across the board. It also accurately predicted Die Back (1645/1656) and Powdery Mildew (1638/1659), with minimal false positives. InceptionV3 (Fig. 6) showed more confusion, particularly between Sooty Mould and Powdery Mildew, misclassifying 70 PM samples as SM and 74 SM as PM, reducing its precision. ResNet152V2 (Fig. 7) performed better overall but misclassified 41 PM as SM and 31 SM as PM. SeResNet152 (Fig. 8) showed improved balance, with high true positives in most classes, including 1650/1651 for CW, and fewer SM–PM confusions. Xception (Fig. 9) struggled the most, misclassifying 70 PM and 70 SM across each other, along with 19 GM and 16 BC errors. Overall, DenseNet201 achieved the most consistent classification performance.

|    | AC   | BC   | CW   | DB   | GM   | HL   | PM   | SM   |
|----|------|------|------|------|------|------|------|------|
| AC | 1621 | 3    | 0    | 1    | 2    | 1    | 0    | 0    |
| BC | 4    | 1634 | 0    | 0    | 6    | 0    | 4    | 4    |
| CW | 0    | 0    | 1647 | 0    | 0    | 0    | 0    | 0    |
| DB | 4    | 0    | 0    | 1645 | 7    | 0    | 0    | 0    |
| GM | 2    | 11   | 0    | 0    | 1625 | 0    | 0    | 4    |
| HL | 14   | 0    | 0    | 0    | 1    | 1647 | 0    | 4    |
| PM | 0    | 0    | 0    | 0    | 1    | 0    | 1638 | 21   |
| SM | 2    | 1    | 0    | 2    | 8    | 1    | 7    | 1615 |

Fig. 5. Confusion matrix of Fine-tuned DenseNet201

|    | AC   | BC   | CW   | DB   | GM   | HL   | PM   | SM   |
|----|------|------|------|------|------|------|------|------|
| AC | 1592 | 5    | 0    | 8    | 13   | 1    | 0    | 3    |
| BC | 7    | 1619 | 0    | 0    | 11   | 0    | 1    | 7    |
| CW | 1    | 0    | 1647 | 0    | 0    | 0    | 0    | 1    |
| DB | 11   | 2    | 0    | 1634 | 12   | 3    | 7    | 4    |
| GM | 29   | 15   | 0    | 0    | 1592 | 1    | 0    | 12   |
| HL | 6    | 0    | 0    | 1    | 8    | 1630 | 3    | 17   |
| PM | 2    | 0    | 0    | 7    | 4    | 1    | 1566 | 74   |
| SM | 1    | 5    | 0    | 0    | 9    | 11   | 70   | 1531 |

Fig. 6. Confusion matrix of Fine-tuned InceptionV3

|    | AC   | BC   | CW   | DB   | GM   | HL   | PM   | SM   |
|----|------|------|------|------|------|------|------|------|
| AC | 1621 | 9    | 0    | 2    | 6    | 1    | 0    | 1    |
| BC | 1    | 1630 | 0    | 0    | 4    | 0    | 8    | 0    |
| CW | 0    | 0    | 1649 | 1    | 0    | 0    | 0    | 0    |
| DB | 2    | 1    | 0    | 1642 | 10   | 0    | 1    | 0    |
| GM | 14   | 5    | 0    | 2    | 1614 | 4    | 0    | 17   |
| HL | 7    | 0    | 0    | 0    | 0    | 1642 | 0    | 8    |
| PM | 0    | 0    | 0    | 1    | 3    | 0    | 1606 | 41   |
| SM | 1    | 4    | 0    | 0    | 12   | 2    | 31   | 1581 |

Fig. 7. Confusion matrix of Fine-tuned ResNet152V2

|    | AC   | BC   | CW   | DB   | GM   | HL   | PM   | SM   |
|----|------|------|------|------|------|------|------|------|
| AC | 1623 | 7    | 0    | 2    | 2    | 0    | 0    | 0    |
| BC | 1    | 1637 | 0    | 0    | 0    | 0    | 0    | 0    |
| CW | 0    | 0    | 1650 | 1    | 0    | 0    | 0    | 0    |
| DB | 4    | 1    | 0    | 1643 | 12   | 0    | 0    | 1    |
| GM | 3    | 2    | 0    | 1    | 1621 | 9    | 0    | 3    |
| HL | 13   | 0    | 0    | 0    | 8    | 1636 | 0    | 1    |
| PM | 1    | 0    | 0    | 1    | 0    | 0    | 1636 | 20   |
| SM | 1    | 0    | 0    | 0    | 5    | 2    | 12   | 1625 |

Fig. 8. Confusion matrix of Fine-tuned SeResNet152

|    | AC   | BC   | CW   | DB   | GM   | HL   | PM   | SM   |
|----|------|------|------|------|------|------|------|------|
| AC | 1615 | 10   | 0    | 1    | 11   | 1    | 0    | 2    |
| BC | 8    | 1603 | 0    | 2    | 7    | 0    | 0    | 5    |
| CW | 0    | 0    | 1648 | 0    | 0    | 0    | 0    | 0    |
| DB | 4    | 4    | 0    | 1635 | 14   | 2    | 4    | 1    |
| GM | 9    | 16   | 0    | 0    | 1590 | 8    | 4    | 1    |
| HL | 10   | 0    | 0    | 1    | 4    | 1634 | 0    | 4    |
| PM | 0    | 0    | 0    | 7    | 2    | 0    | 1570 | 46   |
| SM | 1    | 16   | 0    | 2    | 19   | 3    | 70   | 1590 |

Fig. 9. Confusion matrix of Fine-tuned Xception

### D. Discussion

The experimental results clearly demonstrate that DenseNet201 outperforms the other four evaluated transfer learning models in mango leaf disease classification, achieving the highest overall accuracy of 99.33% and consistently strong class-wise precision, recall, and F1-scores. It performed particularly well in detecting critical disease classes such as Cutting Weevil and Die Back, with perfect precision and recall. ResNet152V2 and SeResNet152 followed closely with accuracies of 99.16%, though they exhibited slightly more misclassifications in visually similar classes like Gall Midge and Healthy. InceptionV3 and Xception, while effective with accuracies above 98%, showed comparatively lower recall in categories such as Sooty Mould and Powdery Mildew, likely due to subtle visual differences. These findings, summarized in the confusion matrices and in Fig. 10, confirm the effectiveness and practical applicability of the fine-tuned DenseNet201 model for precision agriculture, even in resource-constrained environments. Compared to recent studies such as Zhang et al. [9], who achieved 98.42% accuracy with a lightweight CBAM-DBIRNet, and Pathak et al. [13], who reached 99% with a CNN-based Android model, the proposed DenseNet201 approach achieves competitive or superior accuracy and reliability with a relatively simple yet robust design. Its ability to generalize across eight disease classes on real-world orchard data further highlights its potential for practical deployment in smart farming systems.

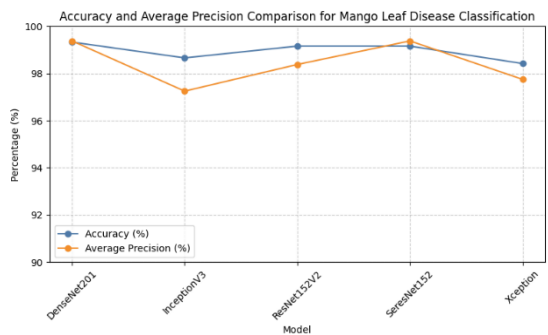

Fig. 10. Accuracy and Precision comparison curve of Fine-tuned models

## V. CONCLUSION

This study demonstrated the effectiveness of fine-tuned transfer learning models for automated mango leaf disease classification, with DenseNet201 outperforming four other CNN architectures. DenseNet201 achieved the highest classification accuracy of 99.33%, along with strong class-wise metrics — precision, recall, and F1-score all reaching up to 100% in critical classes such as Cutting Weevil and Die Back. Compared to ResNet152V2 and SeResNet152 (both ~99.16% accuracy), DenseNet201 exhibited fewer misclassifications, as confirmed by the confusion matrix. InceptionV3 and Xception, while still achieving over 98% accuracy, struggled with visually similar diseases such as Sooty Mould and Powdery Mildew. Training and validation loss curves showed stable convergence without overfitting, and class-level performance was consistently high, confirming DenseNet201's robustness for multi-class disease detection.

Despite these strong results, the study is limited by the dataset's collection from a narrow geographic region and controlled conditions, which may reduce generalizability to real-world orchard environments with diverse lighting, backgrounds, and occlusions. Future work should expand the dataset with more varied conditions, test real-world deployment on mobile or edge devices, and explore lightweight architectures or ensemble techniques to improve efficiency and applicability in resource-constrained settings.